\documentclass[conference]{IEEEtran}
\IEEEoverridecommandlockouts
\usepackage{cite}
\usepackage{amsmath,amssymb,amsfonts,bm}
\usepackage{algorithmic,multirow}
\usepackage{graphicx}
\usepackage{textcomp}
\usepackage{xcolor}
\def\BibTeX{{\rm B\kern-.05em{\sc i\kern-.025em b}\kern-.08em
    T\kern-.1667em\lower.7ex\hbox{E}\kern-.125emX}}
\begin{document}

\title{Masked Cross-image Encoding for Few-shot Segmentation

\thanks{\IEEEauthorrefmark{3}Equal contribution}
}


\author{\IEEEauthorblockN{Wenbo Xu\IEEEauthorrefmark{1}\IEEEauthorrefmark{3}, Huaxi Huang\IEEEauthorrefmark{2}\IEEEauthorrefmark{3}, Ming Cheng\IEEEauthorrefmark{1}, Litao Yu\IEEEauthorrefmark{1}, Qiang Wu\IEEEauthorrefmark{1} and Jian Zhang\IEEEauthorrefmark{1}}
\IEEEauthorblockA{\IEEEauthorrefmark{1}Faculty of Engineering and Information Technology\\
University of Technology Sydney, Sydney, Australia\\
Email: \{Wenbo.Xu@student., Ming.Cheng-3@student., Litao.Yu@, Qiang.Wu@, Jian.Zhang@\}uts.edu.au}
\IEEEauthorblockA{\IEEEauthorrefmark{2}Data61\\Commonwealth Scientific and Industrial Research Organisation, Sydney, Australia\\
Email:huaxi.huang@data61.csiro.au\\
}}
\maketitle

\begin{abstract}
Few-shot segmentation (FSS) is a dense prediction task that aims to infer the pixel-wise labels of unseen classes using only a limited number of annotated images. The key challenge in FSS is to classify the labels of query pixels using class prototypes learned from the few labeled support exemplars. Prior approaches to FSS have typically focused on learning class-wise descriptors independently from support images, thereby ignoring the rich contextual information and mutual dependencies among support-query features. To address this limitation, we propose a joint learning method termed Masked Cross-Image Encoding (MCE), which is designed to capture common visual properties that describe object details and to learn bidirectional inter-image dependencies that enhance feature interaction. MCE is more than a visual representation enrichment module; it also considers cross-image mutual dependencies and implicit guidance. Experiments on FSS benchmarks PASCAL-$5^i$ and COCO-$20^i$ demonstrate the advanced meta-learning ability of the proposed method.
\end{abstract}

\begin{IEEEkeywords}
Few-shot segmentation, prototype learning, cross-image encoding, meta-learning
\end{IEEEkeywords}

\section{Introduction}
Semantic segmentation is a pixel-wise classification problem that aims to predict the class label of each pixel in an image, a task that typically requires a large number of finely labeled images for supervised learning. However, obtaining sufficient image data to train a reliable segmentation model is both expensive and labor-intensive. To address this challenge and reduce the annotation workload, few-shot segmentation methods \cite{RN4, MLC, Part-aware-prototype} have been proposed to learn segmenting previously unseen classes with only one or a few labeled training images, a problem commonly referred to as one-shot or Few-Shot Segmentation (FSS).
\begin{figure}[t]

               \centering

               \includegraphics[scale=0.60]{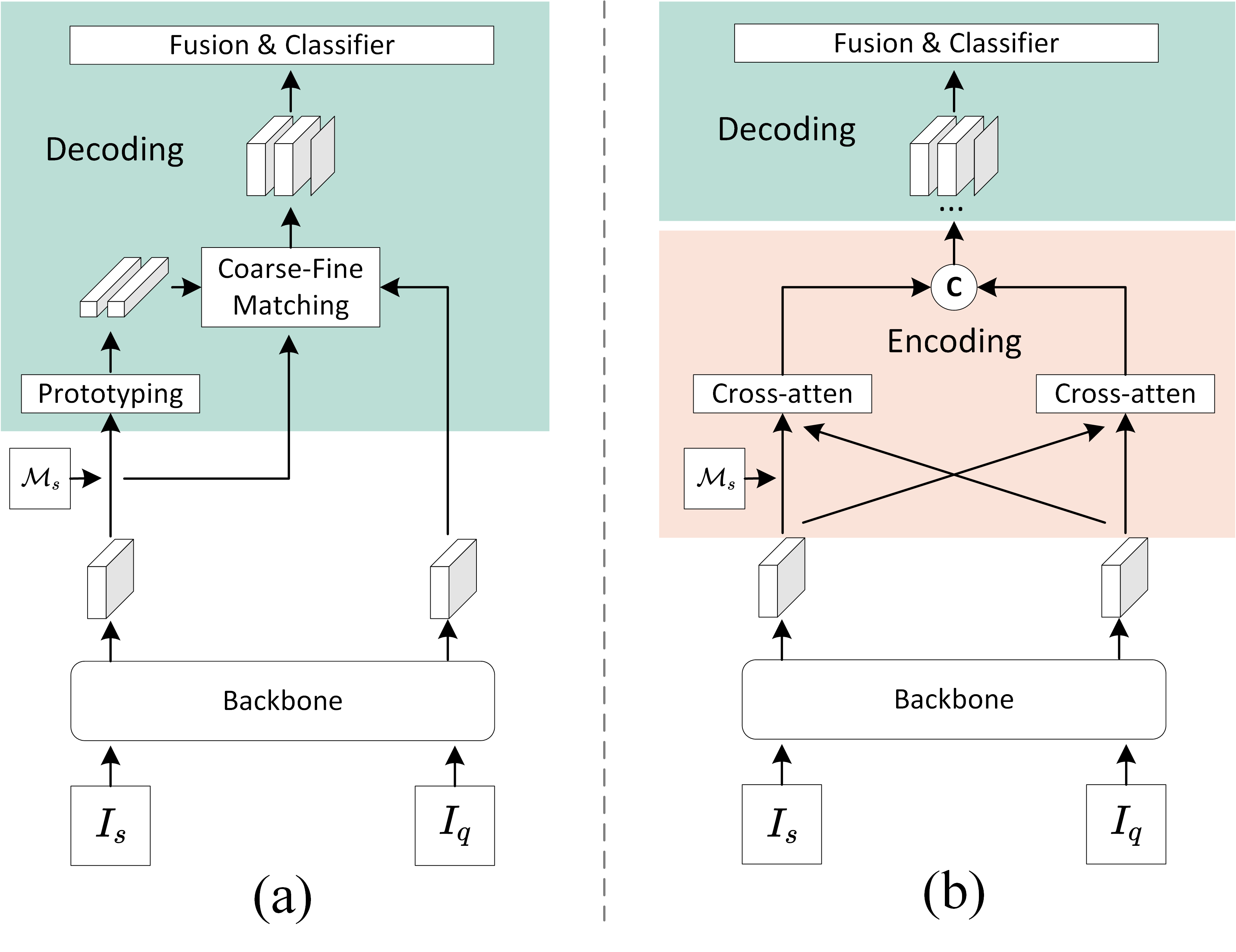}

               \caption{Comparison between conventional ``fixed feature + decoder optimization'' FSS methods and the proposed learnable cross-image feature scheme. (a) Current FSS methods independently learn object prototypes from support features and conduct coarse prototype matching or fine ``pixel-wise'' feature guidance at the decoding stage. (b) The proposed joint encoding scheme makes it possible to learn extra object context from other images before decoding.}

               \label{fig:fig1}

\end{figure}

FSS learning methods generally follow the ``learning to learn'' paradigm, also known as meta-learning \cite{Siamese, tian2020rethinking}, to obtain generalized prototypes that can describe the classes of interest. 
These methods typically apply a two-branch framework~\cite{RN4, RN5,RN10} with a shared frozen pre-trained backbone, where a support branch is used to learn the object prototypes from very few labeled images (support images) and a query branch is utilized to make predictions for a query image on conditioned of the support prototypes. 
Under this framework, masked average pooling (MAP) \cite{ RN5, PANet, RN10} and clustering-based aggregation methods \cite{RN6, ASGNet, Part-aware-prototype} are two popular approaches to learning the representations for support images.While both prototype-based and metric-based approaches have shown promise for few-shot segmentation (FSS), they primarily focus on capturing the overall visual properties of a class and do not effectively incorporate contextual information from a spatial perspective. This can result in a failure to guide features that correspond to the local context of support images, leading to decreased segmentation performance. To overcome this limitation, alignment schemes for FSS employ "features matching" and "decoder optimization" strategies to learn the metric similarities between support and query features extracted from the backbone. Specifically, methods such as \cite{PFENet, ASGNet, MLC} calculate prior maps, or similarity metrics, based on the query and support features to define their correlations. Other methods, including \cite{CWTans, CyCTR, HSNet}, feed support and query features into a decoder (e.g., 4D convolution or transformer) to perform local region matching between query and support images. However, as support and query features are generated independently through a fixed backbone, these alignment methods can still suffer from underdeveloped contextual information mining for support-query pairs that contain the same objects.

It is widely acknowledged that deep learning features extracted from different images of identical objects share similar representations compared to features of other objects. This property makes it more effective to encode an image with features from others that contain the same object when only a limited number of samples are provided. This approach enables better exploration of the semantic representation among features across different image sources and preserves more detailed local contexts that aid in identifying subtle targets. Notably, the FSS two-branch framework naturally provides different image sources containing the same object, namely, the support images and the query image, whose visual features were extracted separately in prior methods. Building on this insight, we observe that self-attention in the Vision Transformer (ViT) \cite{ViT} and cross-attention in \cite{crossattention} can be utilized to capture contextual information of images during token dependency construction. Consequently, we adopt a similar approach to consider joint learning between query and support features for FSS. Specifically, we model cross-image object semantic encoding to identify discriminative local regions, as illustrated in Figure \ref{fig:fig1} (b).

The overall framework of the proposed method is illustrated in Figure \ref{fig:fig2}. The method aims to capture the object semantic mutual relations across support and query images. Unlike CyCTR \cite{CyCTR} which employs Transformer blocks to pass support features to the query decoder, our approach emphasizes the importance of consistent mutual relations between query and support features. To achieve this, we propose a symmetric cross-attention structure called Masked Cross-Image Encoding (MCE), which is designed to assemble bidirectional inter-image relations on multi-level features. The MCE incorporates the support segmentation mask to restrict attention within the localized features of the target objects, thus enhancing the ability to distinguish objects from backgrounds. Additionally, we utilize multi-level features of the support-query images to calculate similarity score matrices, which provide a comprehensive understanding of the correspondence between each position of query features and the support object. Thus, these similarity score matrices help refine the pixel-wise classification accuracy in FSS.

We evaluated the meta-learning ability of our model on two public FSS benchmarks, PASCAL-$5^i$ \cite{VOC} and COCO-$20^i$ \cite{COCO}. The experimental results show that the proposed masked cross-attention encoding helps enrich query features by attending support object regions mutually and, therefore, obtains a strong meta-learning ability, surpassing the prior counterpart methods. In summary, the main contributions of this work are as follows:
\begin{itemize}
\item We propose a masked cross-image encoding method to discover shared visual representations of the target objects in support and query features. By using a symmetric cross-attention structure, MCE can attend to bidirectional inter-image relations on multi-level features, which not only enriches the query features with information from the support object regions but also enhances the support-query interaction, leading to a more favorable meta-learning capability for FSS.
\item We propose to calculate support-query similarity score matrices that reflect the likelihood of a pixel in query features belonging to the foreground. These matrices are then incorporated into our model along with multi-level cross-image features to facilitate final segmentation.
\item Extensive experiments on PASCAL-$5^i$ and COCO-$20^i$ benchmarks under 1-shot and 5-shot settings demonstrate the effectiveness of the proposed MCE and similarity score matrices. The proposed model achieves superior meta-learning performance across all compared state-of-the-art methods. 
\end{itemize}

\section{METHOD}

\subsection{Problem Definition}
In comparison to conventional semantic segmentation, few-shot segmentation involves object classes in the training and testing sets that do not intersect, meaning ${\mathcal{C}_{train}} \cap {\mathcal{C}_{test}} = \emptyset$. This implies that the classes in the testing set are entirely unseen, and no prior knowledge of the testing classes can be obtained during the training stage. From a dataset, we randomly sample several pairs of episodic paradigms, each consisting of a support set $S = \{I^s, M^s\}^k$ and a query image $Q = \{I^q, M^q\}$ that belong to the same class. If there are K annotated images in the support set, the few-shot problem is referred to as K-shot. The objective of few-shot segmentation is to learn a mapping $\mathcal{H_\theta}$ on the training set $D_{train}$ that can accurately predict the query image segmentation mask $M^q$ from the combined input of the support set $S=\{I^s, M^s\}^k$ and query image $I^q$.

\subsection{Model Architecture}

\begin{figure}[t]

               \centering
            \includegraphics[width=8.5cm]{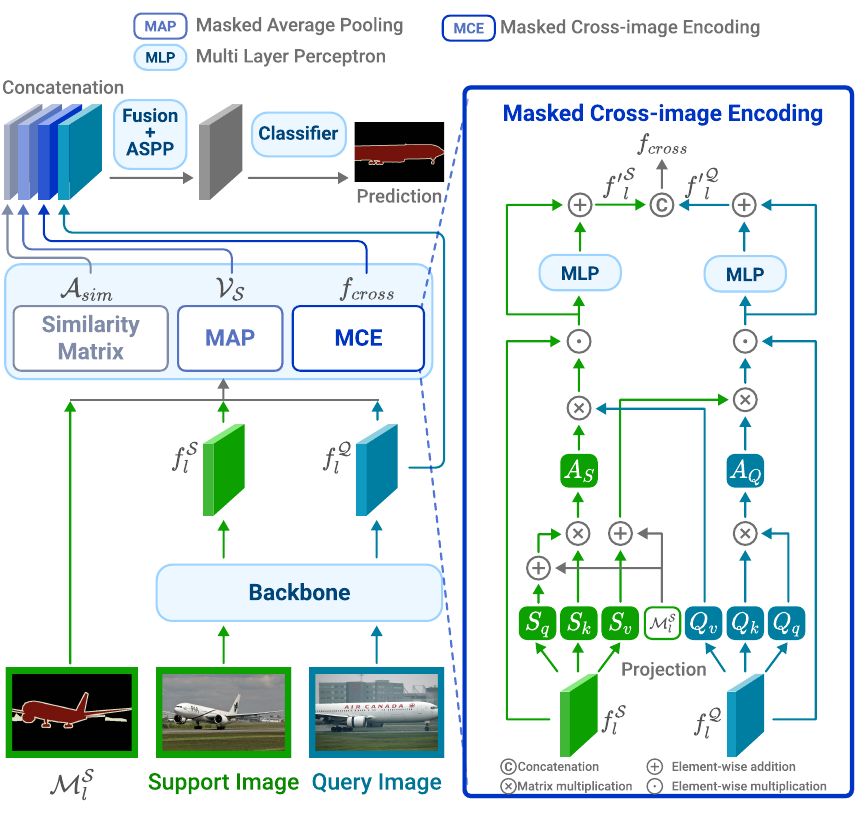}
               \caption{The proposed architecture consists of three distinct modules that take multi-level support-query features $\bm{f}l^S$, $\bm{f}l^Q$ obtained from the backbone network as input along with the support mask to align features. The similarity matrix calculation module evaluates the pixel-wise feature correspondences between the query and support features to derive a similarity score matrix $\mathbf{A}_{sim}$. The Mask Average Pooling (MAP) computes a class-wise prototype $\mathbf{V_s}$ from the support image and corresponding mask. Finally, the Masked Cross-image Encoding (MCE) module leverages the support segmentation mask to confine attention within the localized features of the target objects, thereby improving the ability to differentiate objects from the background as illustrated on the right. These module outputs $\bm{f}_{cross}$, $\mathbf{V_s}$ and $\mathbf{A}_{sim}$, are then concatenated and fused to generate rich features for final prediction.}
               \label{fig:fig2}

\end{figure}

We formulate the overall model architecture in Fig \ref{fig:fig2}. First, multi-level features of a query image and support images are extracted from a pre-trained backbone network. Specifically, the support features extracted from block 2 and 3 and their corresponding masks are utilized to obtain a class-wise prototype vector $\mathbf{V_s}$ by Mask Average Pooling (MAP). The query features and support features extracted from block 4 are utilized to calculate a similarity score matrix $\mathbf{A}_{sim}$. At the same time, all features from the query image and support images extracted from block 2, 3, and 4 are exploited to compute the multi-level cross-image attention map $\bm{f}_{cross}$ through the Masked Cross-image Encoding (MCE) module. The cross-image encoding feature $\bm{f}_{cross}$, the prototype vector $\mathbf{V_s}$, and the similarity score matrix mask $\mathbf{A}_{sim}$, along with query feature $\bm{f}_l^Q$ are concatenated and then fed to the decoder, which is composed of an Atrous Pyramid Pooling (ASPP) module, a $3\times 3$-convolution and a $1\times1$-convolution for binary pixel-wise classification.

The followings present the main components of our cross-attention-based method. We first illustrate the masked attention designed for support-query image pairs. Then, we elaborate on how the symmetric cross-image encoding architecture incorporates features from other images. Finally, we introduce a feature enhancement scheme with feature fusion.
\subsection{Masked Attention Encoding}
The proposed masked attention encoding is inspired by the attention mechanism of the Vision Transformer (ViT) \cite{ViT} model, which first converts an image into a sequence of patches and then linearly maps each patch into tokens with positional embedding. A transformer encoder is composed of a sequence of blocks where each block contains multi-head self-attention (MSA) with a multi-layer perceptron (MLP). Specifically, a scaled dot-product attention is formulated as:
\begin{equation}
\text{Attention} =\text{Softmax}\left(\frac{\mathbf{Q} \mathbf{K}^T}{\sqrt{d}}\right) \mathbf{V},
\end{equation}
where $\mathbf{Q, K, V}$ are the different views of input patch tokens, $d$ is the dimension of each token.

The proposed model adopts a meta-learning scheme with a masked cross-image attention module, which extracts the local features by constraining cross-attention to the foreground region of support features. Specifically, we take multi-level intermediate visual feature representations from support and query images as input. For input feature maps $\bm{f}_l \in \mathbb{R}^ {H_l \times W_l \times C}$, $\mathbf{Q_l, K_l, V_l} \in \mathbb{R}^ {H_lW_l \times N}$ are query-key-value tokens derived from flattened inputs through a projection head $\mathcal{G}_{proj}(\cdot)$, and $H_l$ and $W_l$ are the spatial resolutions of features to attend. The masked attention matrix is computed by:
\begin{equation}
\operatorname{Attention}=\operatorname{Softmax}\left(\left(\bm{\mathcal{M}}_{l}+\mathbf{Q}_l\right) \mathbf{K}_l^{\mathrm{T}}\right) \mathbf{V}_l,
\end{equation}
where the segmentation mask $\mathcal{M}_{l} \in \mathbb{R}^{H_lW_l \times N}$ at feature location $(x,y)$ is calculated by:
\begin{equation}
\bm{\mathcal{M}}_{l}(x, y)= \begin{cases}0 & \text { if } \mathbf{M}_{l}(x, y)=1; \\ -\infty & \text { otherwise. }\end{cases}
\end{equation}

$\mathbf{M}_{l} \in \{0,1\}^{H_lW_l \times N}$ is a binary support mask that transformed from the original image mask $ \mathbf{M}_{l} \in \{0,1\}^{H \times W}$. It is resized to the exact size of $W_l, H_l$ as the input features by linear interpolation, followed by expansion along channel wise and flattened to $\mathbb{R}^{H_lW_l \times N}$.

\subsection{A symmetric cross-image feature encoding method}
The few-shot Siamese segmentation framework\cite{RN4} consists of a support branch and a query branch. The former receives support images with annotated mask learning class representation to guide the query image segmentation in the latter branch. The two-branch guidance architecture has been a source of inspiration for researchers in utilizing Transformer's cross-attention mechanism to enable query features to attend to informative support features, as demonstrated by CyCTR, a closely related work to our study. In CyCTR, support and query features are typically treated as separate entities. The query features are encoded using standard self-attention to obtain a query image representation, which is then unidirectionally passed to a cross-alignment block for query guidance. In contrast, our proposed method suggests that query and support features containing the same class can be integrated as a uniform feature source for generalizing novel class semantic information. By employing cross-image attention, we aim to capture shared representations and semantic relationships among images sharing the same class, which can enhance the meta-learning ability of the few-shot segmentation model.

As it shown in Fig. \ref{fig:fig2}, $l^{th}$ level feature maps $\left(\bm{f}_l^S, \bm{f}_l^Q\right)$ are flattened to sequences of $H_lW_l$ patches, where each patch has $C$ channels. This tokenization can be formulated by $\bm{f_l}=\left[\bm{f}_l^1, \bm{f}_l^2, \ldots, \bm{f}_l^{H_lW_l}\right]$, where $\bm{f}_l^i \in \mathbb{R}^C$. Given a token embedding with mappings $\boldsymbol{W}_S^q, \boldsymbol{W}_S^k, \boldsymbol{W}_S^v$, and $ \boldsymbol{W}_Q^q,  \boldsymbol{W}_Q^k,  \boldsymbol{W}_Q^v$ for support sequence $\bm{f}_S$ and query sequence $\bm{f}_Q$, the query-key-value tokens $\left( \bm{S}_{q}, \bm{S}_k, \bm{S}_v\right)$ and $\left(\bm{Q}_q, \bm{Q}_k, \bm{Q}_v\right)$ can be calculated by:
\begin{equation}
\label{Eq-token}
\left\{\begin{array} { c }
{ \bm{S}_q= \bm{W} _ { S } ^ { q } \bm{f} _ { S } } \\
{ \bm{S}_k = \bm{W}_ { S } ^ { k } \bm{f} _ { S } } \\
{ \bm{S}_v = \bm{W} _ { S } ^ { v } \bm{f} _ { S }}
\end{array} \quad \left\{\begin{array}{c}
\bm{Q}_q=\bm{W}_Q^q \bm{f}_Q \\
\bm{Q}_k=\bm{W}_Q^k \bm{f}_Q \\
\bm{Q}_v=\bm{W}_Q^v \bm{f}_Q
\end{array}\right.\right.
\end{equation}
Note that we omit multi-head attention and multi-level indicator for a concise presentation. We employ the token embedding to preserve the spatial properties, then implement the cross-image encoding by a symmetric cross-attention in two branches: 1) We obtain the support cross feature embedding using the value vectors $\left(\bm{S}_q, \bm{S}_k, \bm{Q}_v\right)$. 2) Similarly, the query cross feature embedding from the query branch is calculated with vectors $\left(\bm{Q}_q, \bm{Q}_k, \bm{S}_v\right)$.

In the support branch, the support embedding first performs self-attention between support tokens ($\bm{S}_q,\bm{S}_k$). Then it conducts cross-attention with a query token $\bm{Q}_v$ to enhance the feature representation of an object. Let $\bm{A}_S \in R^{HW \times HW}$ denote the matrix of self-attention scores obtained via linear mapping:
\begin{equation}
\bm{A}_S = \left(\bm{S}_q+ \bm{\mathcal{M}}\right)\bm{S}_k^{\mathrm{T}},
\end{equation}
where $\bm{S}_q=\left[\bm{S}_q^1,\bm{S}_q^2,...,\bm{S}_q^{HW} \right] \in R^{HW \times C}$ and $S_k=\left[\bm{S}_k^1,\bm{S}_k^2,...,\bm{S}_k^{HW} \right] \in R^{C \times HW}$ are token embeddings to perform self-attention within the query features, which can be obtained using Eq. \ref{Eq-token}. After the self-attention, the model conducts cross-image attention with the attention matrix $\bm{A}_S$ and the query token $\bm{Q}_v$. Moreover, to perform normalization for masked-attention scores and find out the regional semantical relations from the query branch, the scaled attention is calculated as follows:
\begin{equation}
\bm{R}_{S}=\operatorname{softmax}\left(\frac{\bm{A}_S}{\sqrt{d}}\right) \bm{Q}_v,
\end{equation}
where $\bm{Q}_v =\left[\bm{Q}_v^1,\bm{Q}_v^2,...,\bm{Q}_v^{HW} \right] $ is a token of query image features, and $\bm{R}_S \in \mathbb{R} ^{HW \times C}$ represents the masked cross-image feature maps in support branch.

The query cross attention encodes the local semantic information of the support objects into the query feature in a similar way:
\begin{equation}
\bm{A}_Q = \bm{Q}_q \bm{Q}_k^{\mathrm{T}},
\end{equation}

\begin{equation}
\bm{R}_{Q}=\operatorname{Softmax}\left(\frac{\bm{A}_Q}{\sqrt{d}}\right) \left(\bm{S}_v+\bm{\mathcal{M}}\right),
\end{equation}
where the support mask is applied on the support token to remove the background.

The obtained cross-image relation maps $\bm{R}_{Q}$ and $\bm{R}_{S}$ are then fed to the MLP block to further encode the cross-image common information into local regions as: 
\begin{equation}
\left\{\begin{array}{l}
\bm{f}^{\prime S}_{l}= \operatorname{MLP}\left(\operatorname{Norm}\left(\bm{R}_S\right)\right) \\
\bm{f}^{\prime Q}_{l}= \operatorname{MLP}\left(\operatorname{Norm}\left(\bm{R}_{Q}\right)\right) ,
\end{array}\right.
\end{equation}
where Norm represents Layernorm and the MLP block contains of two transformation layers with GELU non-linearity. Finally, the reshaped MLP outputs $\bm{f}^{\prime S}_{l}$ and $\bm{f}^{\prime Q}_{l}$ are aggregated as the cross encoding feature $\bm{f}_{cross}$ by concatenation and $1 \times 1$-convolution.

We notice that the prior map of PFENet \cite{PFENet} determines the query pixel class according to the maximum similarity over the support pixels, which may bias towards a unique support pixel and thus fail to consider the whole object features. Therefore, we propose calculating a mean similarity score matrix to reflect the mean semantic correlation between each query feature position and support object positions. To obtain the relation matrix, we first compute the cosine similarity between every position pair $\left(\bm{x}_q, \bm{x}_s\right)$ from the intermediate masked support feature $\bm{f}_S$ and the query feature $\bm{f}_Q$. The similarity for query pixel and background support pixel pair would be zero, as the mask operation sets background features to zero. The  similarity scores matrix $\bm{A}_{sim} \in \mathbb{R} ^{H_l \times W_l} $ are the mean relation scores as follows:

\begin{equation}
\begin{array}{r}
\bm{A}_{sim}\left(\bm{x}_q, \bm{x}_s\right)=\operatorname{mean}\left(\frac{\bm{x}_q^T \cdot \bm{x}_s}{\left\|\bm{x}_q\right\|\left\|\bm{x}_s\right\|}\right) \\
q \in\left(1,2, \ldots, H_l W_l\right), s \in(1,2, \ldots, H_l W_l),
\end{array}
\end{equation}

The features $\bm{f}_Q, \bm{V}_S, \bm{f}_{cross}, \bm{A}_{sim}$ are concatenated as a whole feature, then it is fed into an ASPP module, where a dilated convolution is used to enlarge the receptive field. Finally, we apply a convolution block and a pixel-wise classifier to predict the final segmentation mask $Pred$:
\begin{equation}
\begin{array}{r}
Pred = \operatorname{Softmax}\left( \operatorname{CLS}\left(\operatorname{Cat\left(\bm{f}_Q, \bm{V}_S, \bm{f}_{cross}, \bm{A}_{sim}\right)}\right)\right),
\end{array}
\end{equation}

Here, $\operatorname{CLS}$ represents a combined operation of an ASPP, an $3\times3$-convolution and a classifier. 

\section{EXPERIMENTS}

\subsection{Dataset and Evaluation Metric}

We evaluated our method on the PASCAL $5^i$ \cite{RN4} and COCO-$20^i$ \cite{COCO} dataset. PASCAL $5^i$ is composed of PASCAL VOC 2012 and extended annotations from SDS \cite{SDS} datasets with 5,953 and 1,449 images for training and validation, respectively. 20 classes were evenly divided into 4 folds $5^i \in{0, 1, 2, 3}$ and each fold contains 5 classes. The dataset COCO-$20^i$ consists of 82,081 training images and 40,137 validation images from 80 object classes divided into 4 folds: $20^i \in{0, 1, 2, 3}$. For the four subsets, three of them were selected as the training set, and the rest one was used as the testing set to validate the effectiveness. Note the training images containing the novel classes on the testing set were removed to prevent information disclosure.

We used the mean intersection over union (mIoU) and foregroud-background IoU (FBIoU) as the evaluation metrics in our experiments.

\begin{table*}[t]

\vspace{-2.0em}

\begin{center}

\caption{Comparison with other FSS networks on PASCAL-$5^i$ under 1-shot and 5-shot settings. BAM$^{*}$ reports the meta-leaner performance.}
\label{table1}

\scalebox{0.90}{

\begin{tabular}{cc|cccccc|cccccc}

\hline

\multirow{2}{*}{Backbone}  & \multirow{2}{*}{Method} & \multicolumn{6}{c|}{1-shot}                           & \multicolumn{6}{c}{5-shot}                            \\ \cline{3-14}

                           &                         & Fold-0 & Fold-1 & Fold-2 & Fold-3 & mIoU\% & FB-IoU\% & Fold-0 & Fold-1 & Fold-2 & Fold-3 & mIoU\% & FB-IoU\% \\ \hline

\multirow{7}{*}{VGG16}     & PANet\cite{PANet}                   & 42.30  & 58.00  & 51.10  & 41.20  & 48.10  & -        & 51.80  & 64.60  & 59.80  & 46.50  & 55.70  & -        \\

                           & FWB\cite{FWB}                    & 47.00  & 59.60  & 52.60  & 48.30  & 51.90  & -        & 50.90  & 62.90  & 56.60  & 50.10  & 55.10  & -        \\

                           & CRNet\cite{CRNet}                  & -      & -      & -      & -      & 55.20  & -        & -      & -      & -      & -      & 58.50  & -        \\

                           & PFENet\cite{PFENet}                  & 56.9   & 68.2   & 54.40  & 52.40  & 58.00  & 72.00    & 59.00  & 69.10  & 54.80  & 52.90  & 59.00  & 72.3     \\

                           & HSNet\cite{HSNet}                   & 59.6   & 65.7   & 59.60  & 54.00  & 59.70  & 73.40    & 64.90  & 69.00  & 64.10  & 58.60  & 64.10  & 76.60    \\

                           & BAM$^{*}$\cite{BAM}                     & 59.90  & 67.51  & 64.93  & 55.72  & 62.02  & -    & 64.02  & 71.51  & 69.39  & 63.55  & 67.12  & -    \\ \cline{2-14}

                           & Ours                     & \textbf{60.59}  & \textbf{69.50}  & \textbf{65.10}  & \textbf{56.30}  & \textbf{62.87}  & \textbf{74.51}    & \textbf{65.63}  & \textbf{72.82}  & \textbf{69.67}  & \textbf{64.70}  & \textbf{68.21}  & \textbf{78.20}    \\ \hline

\multirow{10}{*}{ResNet-50} & PANet\cite{PANet}                   & 44.00  & 57.50  & 50.8   & 44.0   & 49.10  & -        & 55.30  & 67.20  & 61.30  & 53.20  & 59.30  & -        \\

                           & CANet\cite{RN10}                   & 52.50  & 65.90  & 51.30  & 51.90  & 55.40  & -        & 55.50  & 67.80  & 51.90  & 53.20  & 57.10  & -        \\

                           & CRNet\cite{CRNet}                   & -      & -      & -      & -      & 55.70  & -        & -      & -      & -      & -      & 58.80  & -        \\

                           & PPNet\cite{Part-aware-prototype}                   & 48.58  & 60.58  & 55.71  & 46.47  & 52.48  & 69.19    & 58.85  & 68.28  & 66.77  & 57.98  & 62.97  & 75.76    \\

                           & PFENet\cite{PFENet}                  & 61.70  & 69.50  & 55.40  & 56.30  & 60.80  & 73.30    & 63.10  & 70.70  & 55.80  & 57.90  & 61.90  & 73.90    \\

                           & CyCTR\cite{CyCTR}                   & \textbf{67.20}  & 71.10  & 57.60  & 59.00  & 63.70  & -        & \textbf{71.00}  & \textbf{75.00}  & 58.50  & 65.00  & 67.40  & -        \\

                           & HSNet\cite{HSNet}                   & 64.30  & 70.70  & 60.30  & 60.50  & 64.00  & 76.70    & 70.30  & 73.20  & 67.40  & \textbf{67.10}  & 69.50  & 80.60    \\

                           & BAM$^{*}$\cite{BAM}                     & 65.68  & \textbf{71.41}  & 65.56  & 58.93  & 65.40  & -    & 67.28  & 72.38  & 69.16  & 66.25  & 68.77  & -    \\ \cline{2-14}

                           & Ours                     & 65.30  & 71.22  & \textbf{66.17}  & \textbf{61.04}  & \textbf{65.93}  & \textbf{78.10}    & 69.23  & 73.68  & \textbf{70.45}  & 66.76  & \textbf{70.03}  & \textbf{81.33}    \\ \hline

\end{tabular}
}

\end{center}

\end{table*}

\begin{table*}[]

\vspace{-2.0em}

               \caption{Mean IoU(\%) results of 1-shot and 5-shot segmentation on COCO-$20^i$ BAM$^{*}$ reports the meta-leaner performance.} \label{tab:tab2}

               \centering

\label{table 2}

\scalebox{0.90}{

\begin{tabular}{ll|lllll|lllll}

\hline

\multirow{2}{*}{Backbone}  & \multirow{2}{*}{Method} & \multicolumn{5}{c|}{1-shot}                & \multicolumn{5}{c}{5-shot}                 \\ \cline{3-12}

                           &                         & Fold-0 & Fold-1 & Fold-2 & Fold-3 & mIoU\% & Fold-0 & Fold-1 & Fold-2 & Fold-3 & mIoU\% \\ \hline

\multirow{7}{*}{ResNet-50}

                           & ASGNet\cite{ASGNet}                  & -      & -      & -      & -      & 34.56  & -      & -      & -      & -      & 42.48  \\

                           & RePRI\cite{RePRI}                  & 32.00  & 38.70  & 32.70  & 33.10  & 34.10  & 39.30  & 45.40  & 39.70  & 41.80  & 41.60  \\

                           & PPNet\cite{Part-aware-prototype}                   & 28.10  & 30.80  & 29.50  & 27.70  & 29.00  & 39.00  & 40.80  & 37.10  & 37.30  & 38.50  \\

                           & PFENet\cite{PFENet}                  & 36.50  & 38.60  & 34.50  & 33.80  & 35.80  & 36.50  & 43.30  & 37.80  & 38.40  & 39.00  \\

                           & HSNet\cite{HSNet}                   & 36.30  & 43.10  & 38.70  & 38.70   & 39.20  & 43.30  & 51.30  & 48.20  & 45.00  & 46.90  \\

                           & CyCTR\cite{CyCTR}                   & 38.90  & 43.00  & 39.60  & 39.80  & 40.30  & 41.10  & 48.90  & 45.20  & 47.00  & 45.60  \\

                           & BAM$^{*}$\cite{BAM}                     & 41.92  & 45.35  & \textbf{43.86}  & 41.24  & 43.09  & 46.98  & 51.87  & 49.49  & 47.81  & 49.04  \\ \cline{2-12}

                           & Ours                     & \textbf{42.13}  & \textbf{48.30}  & 43.68  & \textbf{42.76}  & \textbf{44.22}  & \textbf{47.80}  & \textbf{55.24}  & \textbf{50.78}  & \textbf{50.33}  & \textbf{51.04}  \\ \hline

\end{tabular}}

\end{table*}
\subsection{Implementation Details}
We used Stochastic Gradient Descent (SGD) as the optimizer, where the momentum and weight decay were set to 0.9 and 0.0001, respectively. Our model was trained for 200 epochs with a base learning rate of 0.0025 and batch size 16 on PASCAL $5^i$. For COCO-$20^i$, models were trained for 100 epochs with a learning rate of 0.005 and batch size 8. In the training stage, we randomly cropped the input images to 473×473. The experiments on PASCAL-$5^i$ were conducted with VGG16 and ResNet-50 backbone, whereas only ResNet-50 was utilized on COCO-$20^i$ dataset. For K-shot setting, the model takes averaged cross-image encoding feature $\{\bm{f}_{cross}^i\}_{i=1}^{K}$, prototype vector  \{$\bm{V}_{s}^i\}_{i=1}^{K}$ and similarity score matrix $ \{\bm{A}_{sim}^i\}_{i=1}^{K}$ to concatenate a fusion feature for final prediction.

\subsection{Results Analysis}

We compared the proposed method with the recent studies on PASCAL-$5^i$ and COCO-$20^i$ under 1-shot and 5-shot settings.

\textbf{Quantitative Results} Table \ref{table1} and Table \ref{table 2} compare the meta-learner performance without filtering background through a base class segmentation network\cite{BAM}. Experimental results show that the proposed method outperforms previous models, achieving 65.93\% and 70.03\% mIoU on PASCAL-$5^i$, 44.22\% and 51.04\% on COCO-$20^i$ under 1-shot and 5-shot settings, respectively. Compared to the Transformer-based method \cite{CyCTR}, our scheme has improved by 2.23\% 1-shot accuracy in terms of mIoU. We can also observe a significant performance improvement of FB-IoU against the model BAM meta-learner\cite{BAM}. On the COCO-$20^i$ dataset, it can be seen that the network with ResNet-50 backbone outperforms all other counterparts by a large margin on the Fold-1 subset, which mostly contributes to the overall mIoU improvement.

\begin{figure}[h]

               \centering
               \includegraphics[scale=1.0]{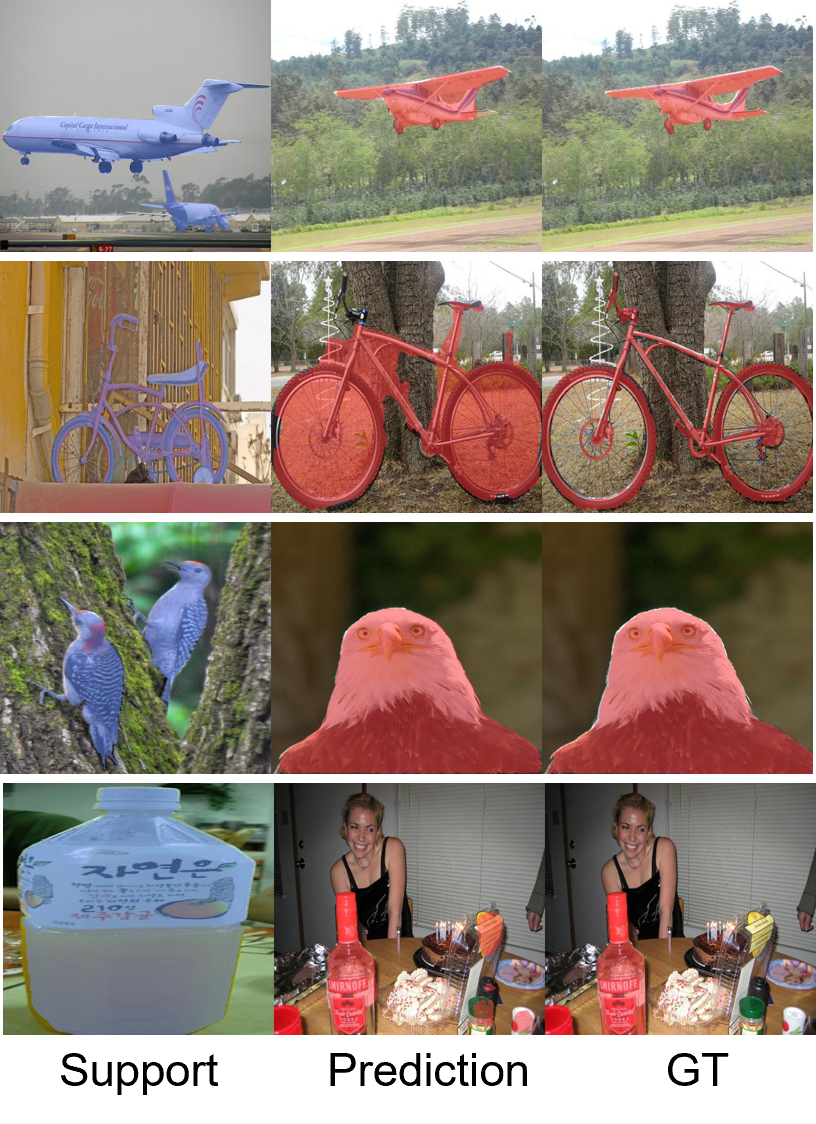}

               \caption{Qualitative results on PASCAL-5$^i$ dataset in 1-shot setting}

               \label{fig:fig4}
\vspace{-1.0em}
\end{figure}

\textbf{Qualitative Results} Fig. \ref{fig:fig4} visualizes some qualitative segmentation results on unseen classes in the 1-shot setting where the first row is the support images with their mask in blue, and the second and third rows are the prediction and ground truth of query images, respectively. It is observed that our method for pixel-wise prediction can accurately cover almost all the target areas with only one support image, demonstrating the benefit of cross-image encoding.

\subsection{Ablation Study}

We first discussed the choices of output maps in the masked cross-image encoding module and then studied the effectiveness of each fused feature. All the tests were conducted under the 1-shot setting on PASCAL-$5^i$ using ResNet-50 backbone.

\textbf{Selection of the MCE outputs}. Due to the nature of the few-shot segmentation task, which mainly focuses on segmenting the query image based on the support images, support features are typically kept constant and used only as references to enhance the query feature. Recent decoder-oriented transformer model\cite{CyCTR} suggests that passing support images that do not correspond to the query mask may negatively impact the self-alignment of the query images. However, it is also worthwhile to enrich the support feature by referencing contextual information from the query feature to reduce underlying inductive bias. To determine the most effective feature in our encoder-oriented symmetric architecture, we separately output the support feature $\bm{f}^{\prime S}_{l}$ and query feature $\bm{f}^{\prime Q}_{l}$ in Figure 2, as well as their fusion feature $\bm{f}_{cross}$, as the final cross-image encoding feature. Experimental results presented in Table 3 indicate that the fused feature outperforms the other two features, suggesting that symmetric cross-image encoding exploits more mutual dependencies than its asymmetric counterparts and leads to implicit feature guidance at the encoding stage.

\textbf{Components Ablations} Table \ref{table 4} presents an analysis of the effectiveness of each component in the proposed network. The results indicate that all proposed modules have a positive impact on performance improvement. Specifically, the absence of cross-image encoding, similarity matrix, and multi-level strategy decreases the prediction mIoU by 0.82\%, 0.32\%, and 1.54\%, respectively, compared to the final aggregation performance. The proposed cross-support-query encoding on multi-level features contributes a noticeable performance gain in enhancing few-shot segmentation performance.

\begin{table}[]

\vspace{-2.0em}

\caption{The performance of optimal output map of the masked cross-image encoding module}
\label{table 3}

\centering
\scalebox{1.0}{
\begin{tabular}{ccc|c}

\hline

$\bm{f}^{\prime Q}_{l}$ & $\bm{f}^{\prime S}_{l}$ & Fusion & mIoU\% \\ \hline

\checkmark   &     &        & 65.24  \\

    & \checkmark    &        & 63.13   \\

    &     & \checkmark       & 65.93  \\ \hline

\end{tabular}
}
\end{table}

\begin{table}[]

\vspace{-1.0em}

\centering

\caption{The results of module performance}
\label{table 4}
\scalebox{1.0}{
\begin{tabular}{ccc|c}

\hline

Cross Map & Sim Mat. & multi-level & mIoU\% \\ \hline

          & \checkmark       & \checkmark           & 64.84  \\

\checkmark         &          & \checkmark           &  65.34      \\

\checkmark         & \checkmark        &             & 64.12  \\

\checkmark         & \checkmark        & \checkmark           & 65.93  \\ \hline

\end{tabular}}
\vspace{-1.0em}
\end{table}

\section{Conclusion}
Our proposed model presents a novel approach to few-shot semantic segmentation by incorporating masked cross-image encoding, mask average pooling, and similarity scores to perform multi-level guidance for FSS. The approach is designed to mine support-query mutual dependencies and introduce a novel approach for jointly encoding shared semantic information and an intuitive scheme for calculating comprehensive pixel-wise relations. The symmetric encoder, utilizing masked cross-image attention, effectively constrains attention within target object regions to enhance support-query feature interaction, which enriches the semantic context of query features and provides implicit guidance for segmentation. Extensive experiments on benchmark datasets demonstrate the effectiveness of our approach, which outperforms compared methods.

\bibliographystyle{IEEEbib}
\bibliography{MCE}

\end{document}